\documentclass{article}

\usepackage{arxiv}

\usepackage[utf8]{inputenc}
\usepackage[T1]{fontenc}
\usepackage{hyperref}
\usepackage{url}
\usepackage{booktabs}
\usepackage{amsmath,amsfonts}
\usepackage{nicefrac}
\usepackage{microtype}
\usepackage{graphicx}
\usepackage{algorithm}
\usepackage{algorithmic}
\usepackage{array}
\usepackage{bm}
\usepackage{multirow}
\usepackage{cite}
\usepackage{enumitem}
\usepackage{balance}

\graphicspath{{./images/}}
\graphicspath{ {./images/} }
\def\BibTeX{{\rm B\kern-.05em{\sc i\kern-.025em b}\kern-.08em
    T\kern-.1667em\lower.7ex\hbox{E}\kern-.125emX}}
\usepackage{balance}

\title{\bf \LARGE CERNet: Class-Embedding Predictive-Coding RNN for Unified Robot Motion, Recognition, and Confidence Estimation}

\author{
 Hiroki Sawada$^{1}$, Alexandre Pitti$^{1}$ and Mathias Quoy$^{1}$%
\thanks{This work was supported by the France Excellence scholarship from the French Ministry for Europe and Foreign Affairs.}
\thanks{$^{1}$Hiroki Sawada, Alexandre Pitti and Mathias Quoy are with Equipe Traitement de l'Information et Systèmes Laboratory (ETIS Laboratory), CNRS UMR8051, CY Cergy-Paris Université, ENSEA, Cergy, France \texttt{hiroki.sawada1@oist.jp}, \texttt{alexandre.pitti@cyu.fr}, \texttt{mathias.quoy@cyu.fr}}%
}

\begin{document}

\maketitle
\thispagestyle{empty}
\pagestyle{empty}

\begin{abstract}
Robots interacting with humans must not only generate learned movements in real-time, but also infer the intent behind observed behaviors and estimate the confidence of their own inferences.
This paper proposes a unified model that achieves all three capabilities within a single hierarchical predictive-coding recurrent neural network (PC-RNN) equipped with a class embedding vector, CERNet, which leverages a dynamically updated class embedding vector to unify motor generation and recognition.
The model operates in two modes: generation and inference.
In the generation mode, the class embedding constrains the hidden state dynamics to a class-specific subspace; in the inference mode, it is optimized online to minimize prediction error, enabling real-time recognition.
Validated on a humanoid robot across 26 kinesthetically taught alphabets, our hierarchical model achieves 76\% lower trajectory reproduction error than a parameter-matched single-layer baseline, maintains motion fidelity under external perturbations, and infers the demonstrated trajectory class online with 68\% Top-1 and 81\% Top-2 accuracy.
Furthermore, internal prediction errors naturally reflect the model's confidence in its recognition.
This integration of robust generation, real-time recognition, and intrinsic uncertainty estimation within a compact PC-RNN framework offers a compact and extensible approach to motor memory in physical robots, with potential applications in intent-sensitive human–robot collaboration.
\end{abstract}


\section{Introduction}

Robots that share workspaces and interact with humans must be capable of \emph{generating} learnt behaviors, \emph{recognizing} human intentions, and \emph{estimating their own confidence} in uncertain, real-time environments.
Predictive–Coding (PC) networks offer a promising framework for achieving such capabilities by minimizing prediction errors through top-down prediction and bottom-up inference.
However, existing implementations often treat these functions separately, leaving few PC-based models that unify all three processes within a single closed-loop architecture for robotics.
Even among those few, such integration typically relies on complex, multi-module systems; by contrast, our approach implements all functions within a single PC recurrent neural network (RNN).

While various predictive-coding networks have been proposed for visual prediction~\cite{lotter2016deep}, visuo-motor control~\cite{tani2019accounting}, or hierarchical perception~\cite{qiu2025deep}, most still rely on external classifiers or post-hoc thresholding for recognition and confidence estimation.  
Moreover, emerging methods that embed confidence-related dynamics within inference~\cite{granier2024confidence,sharafeldin2024active} often require separate decision modules and remain unvalidated on physical robotic platforms.
To concretely evaluate these challenges, we focus on the task of learning alphabet writing trajectories and recognizing the corresponding character classes from observed trajectory segments in real time.  
Within this setting, a single, parameter-efficient, multi-layer predictive-coding RNN that unifies action generation, label inference, and intrinsic confidence estimation on physical hardware under external disturbances is still lacking.  
To fill this gap, we introduce the Class-Embedding predictive-coding Recurrent NETwork (CERNet).

To jointly achieve unified action generation, label inference, and confidence estimation under a single model, we combine three design principles in CERNet.
First, a multi-layer architecture allows higher layers to maintain abstract motion intentions across longer timescales \cite{yamashita2008emergence}, enabling the model to stably reproduce demonstrated trajectories even on physical hardware.
In our experiments, this structure outperforms single-layer networks of equivalent parameter size, which fail to accurately reproduce the demonstrated trajectories under real-world conditions.
Second, predictive-coding framework continuously minimizes prediction errors between top-down expectations and sensory input \cite{rao1999predictive}.
These dynamics enable the model to adapt to external perturbations (e.g., unforeseen forces or deviations arising in real-world robotic tasks) during execution, recovering towards the target trajectory by updating internal states and correcting future predictions in real time.
Third, we introduce a class embedding vector that is updated online via the same prediction-error minimization.
As the network observes a motion, the embedding vector gradually drifts toward the corresponding latent subspace, effectively serving as a self-organizing key for label inference. 
Moreover, by analyzing reconstruction errors of past observations, the network can internally estimate the confidence of its recognition, without requiring any explicit classifier or output module.
While each of these principles has been studied in isolation or in simulation, this is the first work to integrate them into a single predictive-coding RNN and validate their combined effectiveness on a physical robotic platform.

Therefore, our experiments reveal four key capabilities of unified architecture of CERNet. 
First, its hierarchical structure greatly reduces reproduction error, highlighting the benefit of layered temporal abstraction. 
Second, the predictive-coding framework confers robustness to disturbances: after perturbations, the robot autonomously recovers to the intended trajectory. 
Third, by tuning the prediction–recognition balance, the model can infer the intended motion class in real time with high accuracy. 
Fourth, the same internal error signal that guides inference also indicates confidence, enabling implicit self-evaluation without extra classifiers. 
These results demonstrate that predictive-coding architectures can serve not only as robust motor generators but also as flexible intention recognizers within a single compact model, providing a foundation for future applications in human–robot collaboration.

The remainder of this paper is organized as follows: Section 2 reviews related work, Section 3 introduces the proposed CERNet architecture, Section 4 describes the experimental setup with the humanoid robot Reachy, Section 5 presents results from both simulation and real-world trials, and Section 6 concludes with a summary and future directions.

\section{Related Works}

\subsection{Unifying Recognition, Generation and Confidence}

Many recent robot learning models focus either on \emph{recognition}, inferring task identity or intention from sensory observations, or on \emph{generation}, producing learned trajectories from latent representations.  
Most frameworks treat these two functions separately, using distinct architectures for perception and motion control.  

For recognition, common approaches include discriminative RNNs trained with cross-entropy loss~\cite{duan2017one}, latent-variable inference via modular policies~\cite{devin2017learning}.
These methods typically require large labeled datasets and assume segmented, unambiguous observations.
Generation, on the other hand, is often addressed with autoregressive sequence models such as LSTMs~\cite{merel2018neural}, Transformers~\cite{vaswani2017attention}, and more recently, policy-conditioned diffusion models~\cite{chi2023diffusion}.  
While these models achieve high-fidelity generation, they generally lack the capacity for real-time task inference.
Only a few works attempt to unify these roles within a single architecture.  
Latent-variable models such as GAIL~\cite{ho2016generative}, World Models~\cite{ha2018recurrent}, or MTRNN-based frameworks~\cite{yamashita2008emergence} leverage shared internal dynamics for both recognition and action generation.  
However, these typically rely on offline clustering or external classifiers and are rarely evaluated on real-time, closed-loop robotic platforms.

While a few studies have explored confidence estimation alongside recognition or generation, they typically rely on separate modules or post-hoc uncertainty measures.  
While recent studies embed confidence-related dynamics within inference, such as precision-weighted hierarchies \cite{granier2024confidence} and uncertainty-minimizing active sensing \cite{sharafeldin2024active}, they often rely on separate decision stages and remain unvalidated on physical hardware.
In contrast, our approach leverages the same internal prediction error that drives generation and recognition, allowing confidence to emerge as an intrinsic property of the network’s ongoing inference process.

Our work addresses this gap by demonstrating that CERNet can serve simultaneously as a generator, a recognizer, and an estimator of its own confidence.  
The same network dynamics update online through prediction-error minimization, allowing recognition to emerge gradually as motion is observed, confidence to be estimated from reconstruction error, and generation to take over seamlessly without retraining or architectural changes.

\subsection{Predictive Coding RNNs}

Predictive-coding recurrent neural networks (PC-RNNs) maintain internal representations by alternately generating top-down predictions and correcting them using bottom-up inference~\cite{rao1999predictive}.  
This recursive process not only enables online adaptation but also naturally couples generation and recognition: actions are generated by minimizing prediction error, while recognition emerges through internal state updates that best explain sensory observations.

A key insight of the PC framework is that the magnitude and direction of internal prediction errors encode information about the plausibility of the current internal belief~\cite{friston2009predictive}.  
In other words, the network's confidence can be inferred directly from its internal error dynamics, without requiring separate uncertainty estimation modules.
This property makes PC-RNNs particularly attractive for real-time robotics, where decisions must be made under uncertainty and with limited computation.  
Furthermore, when combined with a class embedding vector that is updated through the same prediction-error minimization loop, the model can unify label inference and motor generation within a single dynamical system.  
The embedding trajectory gradually converges to a class-specific latent subspace, guiding both inference and generation, while reconstruction error implicitly reflects recognition confidence.

Existing PC-RNN variants, such as PV-RNN~\cite{ahmadi2019novel} or Active Predictive Coding~\cite{rao2024sensory}, have explored stochastic representations and multi-timescale dynamics, respectively.  
Class-gated extensions of PC-RNNs have also been proposed to mitigate pattern interference and enable continual sequence learning~\cite{annabi2021bidirectional,annabi2022continual}.  
Complementary work on Free-Energy–based training schemes, including INFERNO and its gated or spiking variants~\cite{pitti2017iterative,pitti2020gated}, demonstrates how iterative error minimization can scale to long temporal structures.  
However, these studies still separate recognition from generation via offline decoding or manual gating, and rarely exploit prediction-error magnitudes as a direct confidence signal.  
We therefore start from the single-layer class-gated PC-RNN of~\cite{annabi2021bidirectional} and expand it in four directions: a three-layer hierarchy, deployment on real robotic hardware, real-time class inference, and intrinsic confidence estimation from prediction errors.  
These extensions turn a simulation-only proof-of-concept into a unified, closed-loop system capable of robust generation, recognition, and self-evaluation on a physical platform.

\section{Methods}

\subsection{CERNet: Class-Embedding Predictive-Coding RNN}
\label{sec:past_reconstruction}

\begin{figure}[htbp]
  \centering
  \includegraphics[width=0.95\linewidth]{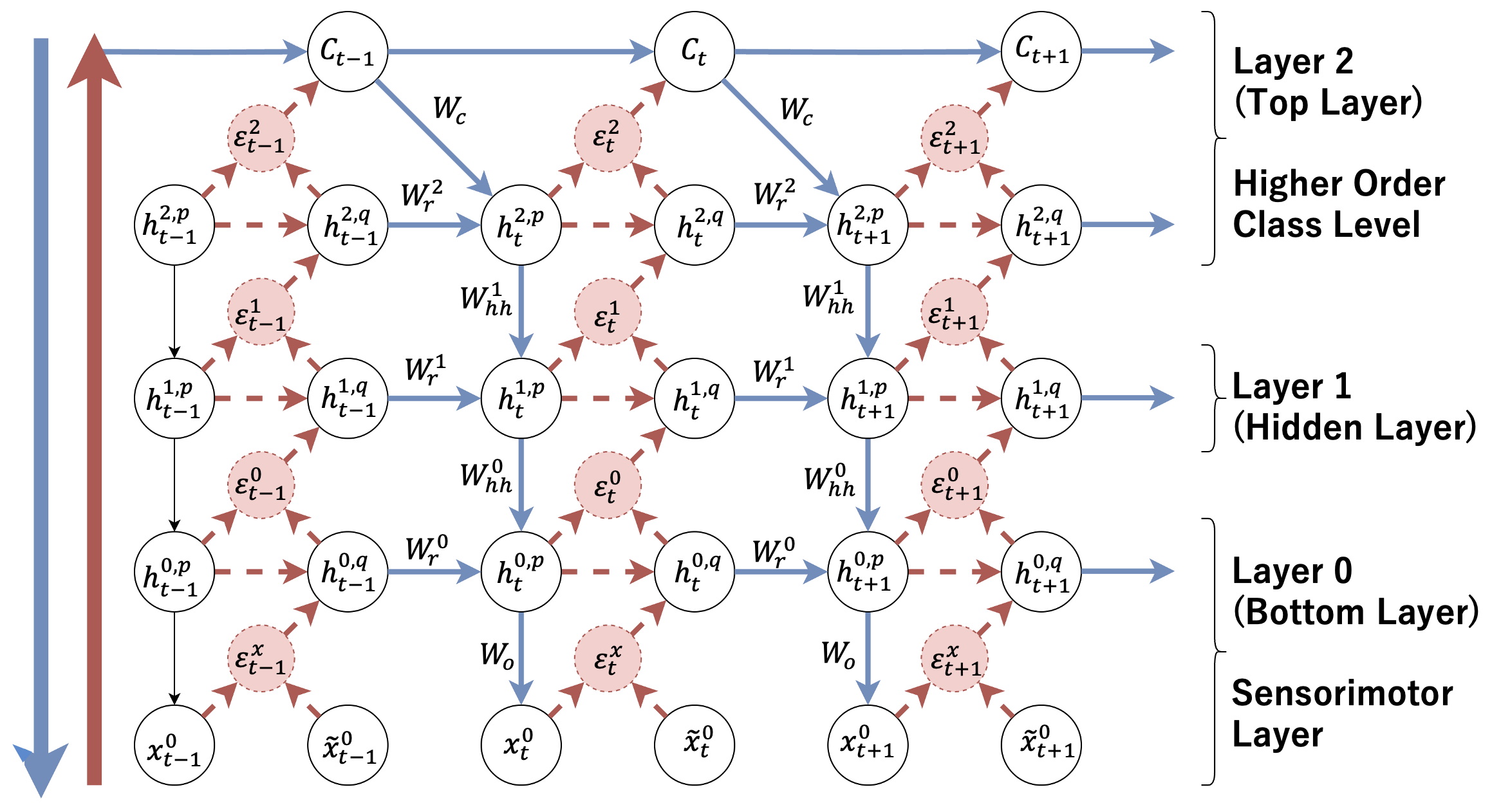}
  \caption{Schematic illustration of CERNet. 
  The model integrates top-down predictions and bottom-up errors across hidden states in multiple layers.
  Blue and red arrows indicate computation for forward propagation and error propagation, respectively.
  Refer to Eqs.~\ref{eq:ClassEmbeddingVector}--\ref{eq:adam_clip} for detailed explanation of each variable.}
  \label{fig:hpc_rnn_schematic}
\end{figure}

To realize unified generation, recognition, and confidence estimation under a single dynamical system, we propose the Class-Embedding Predictive-Coding RNN (CERNet).  
Its architecture extends conventional predictive-coding RNNs by incorporating a class embedding vector that dynamically modulates the internal dynamics via the same prediction-error minimization loop.
The model's overall structure is shown in Fig. \ref{fig:hpc_rnn_schematic}.
Firstly, it is equipped with a one-hot-like class embedding vector \( \mathbf{C} \in \mathbb{R}^K \), where \( K \) denotes the number of predefined classes. The vector \( \mathbf{C}^k \) of selected class index \( k \in \{1, \dots, K\} \) is defined as:
\begin{equation}
    C_i^k =
    \begin{cases}
    1, & \text{if } i = k \\
    0, & \text{otherwise}
    \end{cases}
    \quad \text{for } i = 1, \dots, K.
    \label{eq:ClassEmbeddingVector}
\end{equation}
In other words, only the \( k \)-th element of \( \mathbf{C} \) is 1, and all others are 0.

This supports the network in distinguishing different sequential patterns in the training dataset and generating each trained pattern.
Secondly, each internal state of CERNet consists of a pair of prior and posterior values, each of which corresponds to the internal state before and after observation, respectively.
Throughout the operation of this model, internal states are influenced by the discrepancy between prediction and observation, and updated to minimize those discrepancies.

During forward computation of CERNet, the prior $p$ internal state $h_{t}^{n, p}$ at time $t$ and layer $n$ is updated based on the recurrent dynamics, top-down inputs (except for the top layer), and the cause input (for the top layer only), as follows:

\begin{align}
    h_{t}^{n, p} &= \left( 1 - \frac{1}{\tau_h^n} \right) h_{t-1}^{n, q} \nonumber \\
    &\quad + \frac{1}{\tau_h^n} \left(
        \tanh(h_{t-1}^{n, q}) W_r^{n\top}
        + \delta_{n = N} C W_c^{\top} \right. \nonumber \\
    &\qquad + \delta_{n < N} \tanh(h_t^{n+1, q}) W_{hh}^{n\top} \left. 
        + b_r^n
    \right),
    \label{eq:forward_h}
\end{align}
where $\tau_h^n$ is a time constant controlling the update rate at layer $n$.
$W_r^{n}$ and $W_{hh}^{n}$ are the recurrent and top-down weight matrices, respectively, and $b_r^n$ is a bias term.
$\delta_{n < N}$ and $\delta_{n = N}$ are indicator functions that activate the corresponding term only when the condition is satisfied.
$C$ is the class-embedding vector, applied only in the top layer ($n = N$).

Prediction at time $t$, $x_t^0$, is then calculated in the bottom layer as follows:

\begin{equation}
    x_t^0 = \tanh{(h_t^{0, p})} W_o^T + b_o,
    \label{eq:forward_output}
\end{equation}
where $h_t^{0,p}$ is the corresponding prior hidden state obtained from Eq. \ref{eq:forward_h}.
$W_o$ and $b_o$ denote the output‐layer weight matrix and bias vector, respectively.

After the model computed the prediction for time $t$, it is fed with the observation $\tilde{x}_t^o$.
Utilizing the bottom-up error between the prediction and the observation $\varepsilon_t$ ($=x_t^o - \tilde{x}_t^o$), the posterior $q$ internal state $h_t^{n, q}$ deviates from the prior internal state to minimize the bottom-up error:

\begin{equation}
    S_t^{n} \;=\;
    \begin{cases}
        \varepsilon_t\, W_o, & n = 0,\\[4pt]
        \varepsilon_t^{\,n-1}\, W_{hh}^{\,n-1}, & n > 0.
    \end{cases}
    \label{eq:def_alpha_S}
\end{equation}

\begin{equation}
    h_{t}^{\,n,q}
    \;=\;
    h_{t}^{\,n,p}
    \;-\;
    \alpha_{h}^n\,
    \bigl[1-\tanh^{2}\!\bigl(h_{t}^{\,n,p}\bigr)\bigr]\;
    S_t^{n},
    \label{eq:update_hq}
\end{equation}

\begin{equation}
    \varepsilon_t^{\,n} = h_{t}^{\,n,p} - h_{t}^{\,n,q},
    \qquad
    \varepsilon_t      = x_t^{0}      - x_t^{\mathrm{obs}},
    \label{eq:error_defs}
\end{equation}

\noindent
where $\alpha_{h}^n$ represents the learning-rate parameter for each layer, and $S_t^{n}$ provides the appropriate error-driven—sensory prediction error at $n{=}0$ (bottom layer) or hierarchical prediction error from the layer below for $n{>}0$ (top layer).
Layer-wise prediction errors $\varepsilon_t^{\,n}$ are recursively fed upward via $S_t^{n+1}$, while $\varepsilon_t$ denotes the sensory prediction error at the bottom layer.

The proposed CERNet operates under a unified predictive-coding framework across three functional phases:
1) a \emph{training phase} for learning from labeled demonstrations,  
2) a \emph{generative phase} for autonomous reproduction of motion, and
3) an \emph{inference phase} for online recognition of task identity.
All three phases share the same network architecture and internal dynamics, relying on the minimization of prediction error to update both hidden states and, when applicable, the class embedding vector $\mathbf{C}$.

During the \emph{training} phase, the network receives labeled demonstrations: the observation sequence (end-effector trajectories) together with the corresponding class identifier; the model learns its parameters.
After training, in the \emph{generation} phase, a target class identifier is specified and the network autonomously reproduces the motion in closed loop on the robot; internal states are updated online during execution, but the learned parameters and the class representation remain fixed.
In the \emph{inference} phase, the observed end-effector trajectory is provided incrementally from the robot; the model updates its internal states online and \emph{infers} the class identity from the partial observation.

During the \emph{training phase}, after computing the forward pass, the model parameters, $\theta=\{W_o,b_o,W_{c},W_{hh}^{\,n},W_{r}^{\,n},b_{r}^{\,n}\}$ are optimised by back-propagation to minimise the \emph{prediction-error loss} (equivalent, up to a constant factor, to the variational free energy):
\begin{equation}
    \mathcal{L}_{\mathrm{PE}}
         =\frac{1}{2}\sum_{t}\!
          \Bigl(\lVert\varepsilon_t\rVert_2^2 +
                \sum_{n=0}^{N}\alpha_n\,
                \lVert\varepsilon_t^{\,n}\rVert_2^2\Bigr),
    \label{eq:prediction_error_loss}
\end{equation}
while the variational free-energy (VFE) is computed in parallel only for logging and analysis.
All layer-specific gradients are concatenated into a single vector
\begin{equation}
    g_t \;=\;
    \Bigl[
    \frac{\partial\mathcal{L}}{\partial W_o},\,
    \frac{\partial\mathcal{L}}{\partial b_o},\,
    \frac{\partial\mathcal{L}}{\partial W_c},\,
    \frac{\partial\mathcal{L}}{\partial W_{hh}^{\,0{:}N-1}},\,
    \frac{\partial\mathcal{L}}{\partial W_{r}^{\,0{:}N}},\,
    \frac{\partial\mathcal{L}}{\partial b_{r}^{\,0{:}N}}
    \Bigr],
    \label{eq:grad_vec}
\end{equation}
which is subjected to global $L_2$-norm clipping ($G_{\max}=1.0$) and updated with Adam,
\begin{equation}
    \!\!
    \begin{aligned}
    \hat g_t &= g_t\;\cdot\!
               \min\!\Bigl(1,\frac{G_{\max}}{\lVert g_t\rVert_2+\varepsilon}\Bigr), 
            m_t = \beta_1 m_{t-1} + (1-\beta_1)\hat g_t,\;\; \\
        v_t &= \beta_2 v_{t-1} + (1-\beta_2)\hat g_t^{\,2},
            \theta_t = \theta_{t-1} -
               \eta\,
               \dfrac{m_t/(1-\beta_1^{\,t})}
                     {\sqrt{v_t/(1-\beta_2^{\,t})}+\varepsilon},
    \end{aligned}
    \label{eq:adam_clip}
\end{equation}
with $\beta_1{=}0.9$, $\beta_2{=}0.999$, and $\varepsilon{=}10^{-6}$.

During both the \emph{generation phase} and the \emph{inference phase}, the model no longer updates its network weights.
Instead, the internal hidden states $h_t^n$ are dynamically updated at each timestep to minimize prediction errors, using the same error-driven update rules as in the training phase.
Specifically, Eqs.~\eqref{eq:def_alpha_S}–\eqref{eq:error_defs} govern the propagation of prediction errors and the corresponding updates to each layer’s internal state, thereby enabling the model to adapt its predictions online without parameter changes.

In addition to updating $h$, the \emph{inference phase} further refines the class embedding vector $\mathbf{C}$.
This is achieved through a procedure known as \textbf{past reconstruction error minimization}, inspired by inference techniques in stochastic PC models such as PV-RNN~\cite{ahmadi2019novel}, where internal states are optimized over past trajectories.
Rather than relying solely on the current timestep, $\mathbf{C}$ is iteratively updated to minimize the accumulated prediction error over the full observed sequence up to time $t$.
While the network weights remain fixed, the embedding vector is refined using gradient descent:

\begin{equation}
    \mathbf{C} \leftarrow \mathbf{C}
    - \alpha_c \cdot
    \frac{1}{\,t - (t-W)_{+} + 1\,}
    \sum_{\tau=(t-W)_{+}}^{t}
    \nabla_{\mathbf{C}} \mathcal{L}_{\mathrm{PE}}(\tau),
\end{equation}

with the gradients $\nabla_{\mathbf{C}} \mathcal{L}_{\text{PE}}(\tau)$ back-propagated through the hierarchical PC layers using the same equations as above.
To reduce computational cost, the gradients are averaged only over a sliding window of length $W$, i.e., 
$\tau$ runs from $(t-W)_{+}$ to $t$ with normalization by the number of terms, where $(\cdot)_{+}=\max(\cdot,0)$.

This inference–update cycle is repeated for $n_{\text{iter}}$ iterations at each timestep, where the current class embedding $\mathbf{C}$ is used to regenerate a prediction sequence $x_{0:t}^{\text{pred}}$, and then updated to reduce its reconstruction error.
Through this iterative process, $\mathbf{C}$ gradually drifts toward the latent subspace most consistent with the observed sequence.
We refer to this approach as \textbf{past reconstruction}, which unifies recognition and generation within the same predictive-coding framework.

\subsection{Humanoid Robot: Reachy}

All physical experiments were conducted on Reachy 2021\footnote{Pollen Robotics: \url{https://pollen-robotics.github.io/reachy-2021-docs/}}, a humanoid robotic platform developed by Pollen Robotics.  
In this study, we used the left arm (7 DoF), which includes Dynamixel motors controlled via the Reachy SDK over a UDP interface.
Joint commands were sent using the minimum-jerk interpolation mode provided by the \texttt{reachy-sdk}, with a control frequency of $20$ Hz.  
The end-effector position (3-dimensional Cartesian) was computed using the built-in forward/inverse kinematics solvers, and all training trajectories were recorded through kinaesthetic teaching.  
During the interaction phase, the predicted end-effector trajectories were converted into joint angles via inverse kinematics and executed directly on the robot.
To ensure consistent execution, each trajectory was preceded by a reset-to-home motion.  
No external sensor (e.g., camera or IMU) was used in this study.

\section{Experiment}

All experiments were conducted as described below.
To promote transparency and reproducibility, \textbf{all raw trajectory data, model weights, training scripts, and analysis code are publicly available}.%
\footnote{\url{https://git.cyu.fr/hiroki/cernet_paperresultreproduction} (eCILL license).}

\subsection{Experimental Setup}

\subsubsection{Train Data-Set Preparation}

To collect training data, we physically guided the left arm of the humanoid robot Reachy to draw each letter of the English alphabet on a table.
The corresponding end-effector positions were estimated using Reachy's built-in inverse kinematics and recorded at a sampling rate of 20 Hz.
Each sequence was recorded for 100 time steps without interpolation.
After completing each character, we held the end-effector stationary for the remaining steps to preserve a consistent sequence length.
The final dataset thus comprises 26 distinct trajectory sequences, each representing one alphabet character, and each sequence consists of 100 time steps.

\subsubsection{Network Configuration}

We prepared six CERNet variants with varying hidden sizes and depths, as summarized in Table~\ref{tab:model-configs}.  
These include three single-layered models (A:~\textbf{SingleMini}, B:~\textbf{SingleStandard}, C:~\textbf{SingleLarge}) and three hierarchical models (D:~\textbf{MultiMini}, E:~\textbf{MultiStandard}, F:~\textbf{MultiLarge}), each differing in total parameter count and node allocation.

This setup allows us to compare the effectiveness of the hierarchical architecture against single-layered models under similar model capacities.
To ensure fair comparison, the number of parameters in each pair of models—Mini, Standard, and Large—has been adjusted so that the single-layered and hierarchical models have approximately the same total number of parameters.
To guarantee statistical reliability in the analysis of experimental results, we have trained each model 10 times with identical parameters (Table \ref{tab:model-configs}).
\# of nodes, $\tau_h$, $\alpha_h$, \# of params indicate the number of nodes in each hidden layer (with larger sizes in the top layer and smaller ones in the bottom layer for hierarchical models), the time constant, the update rates of posterior hidden state in each hidden layer and the number of trainable parameters, respectively.

\begin{table}[h]
    \centering
    \caption{Model configurations for each CERNets}
    \label{tab:model-configs}
    \resizebox{\linewidth}{!}{%
    \begin{tabular}{l l llll}
        \hline
        Type & ID & Nodes & $\tau_h$ & $\alpha_h$ ($\times 10^{-2}$) & Params \\
        \hline
        \multirow{3}{*}{Single} 
            & A) & [50]   & [10]         & [1]              & 3.9k   \\
            & B) & [150]  & [10]         & [1]              & 22.9k  \\
            & C) & [300]  & [10]         & [1]              & 90.9k  \\
        \hline
        \multirow{3}{*}{Multi}
            & D) & [50,15,7]     & [10,20,40] & [1, 0.05, 0.0005] & 4.6k   \\
            & E) & [120,40,20]   & [10,20,40] & [1, 0.05, 0.0005] & 22.9k  \\
            & F) & [250,70,20]   & [10,20,40] & [1, 0.05, 0.0005] & 90.9k  \\
        \hline
    \end{tabular}%
    }
\end{table}

Each model was trained for 10,000 epochs to minimize the \emph{prediction-error loss} shown in Eq. \ref{eq:prediction_error_loss}.
The network weights were initialized randomly, using a different random seed for each training run.

\subsubsection{Alphabet Drawing Experiment}
\label{sec:AlphabetDrawingExperiment}

After training, the model parameters loaded for evaluation were taken from the epoch with the lowest loss recorded during training.
All 60 trained network instances were first evaluated in simulation to verify closed-loop prediction stability across the full set of 26 alphabet characters.  
Based on this screening, four models—\textbf{MultiStandard}, \textbf{MultiLarge}, \textbf{SingleStandard}, and \textbf{SingleLarge}—were selected for physical experiments on the humanoid robot \textit{Reachy}, as they demonstrated sufficiently stable and safe behavior in the simulated loop.

In the real-world closed-loop interaction setting, the network produced a 3D Cartesian prediction of the left-arm end-effector position at each control step.
This prediction was forwarded to Reachy’s built-in impedance controller, which converted it to joint angle commands via inverse kinematics and moved the arm accordingly.
The controller then returned the measured end-effector position to the network, closing the prediction–observation loop.
This cycle was repeated for 100 consecutive timesteps while the robot traced each of the 26 alphabet characters learned during training.
The class embedding vector was initialized to the one-hot vector corresponding to the correct character for each trial.
The drawn trajectory was then evaluated by calculating the Dynamic Time Warping (DTW) score, a distance-based measure that quantifies similarity between two temporal sequences by optimally aligning them even if they vary in speed or timing, with respect to the training data.
Throughout the experiment, the round-trip latency between the host computer and the robot remained below 30 ms.


\subsubsection{Perturbation Resistance Experiment Using Reachy}

In addition to the closed-loop interaction setting, we have programatically inserted an external perturbation into the observation of the end-effector position during the experiment for robustness evaluation.
Specifically, while Reachy was operated by a trained network, we applied an external disturbance from time step 40 to 45, which forced its arm to drift from the predicted trajectory.
We then evaluated the ability of the model to recover and return to the learned trajectory.
This perturbation experiment was conducted only on model F: \textbf{MultiLarge}, as other models did not consistently demonstrate stable closed-loop behavior on the physical platform and were deemed unsuitable for safe evaluation, based on the results of the alphabet drawing experiment in Section~\ref{sec:AlphabetDrawingExperiment}.

\subsubsection{Class Inference \& Confidence Estimation Experiment}

In a separate experiment, we evaluated the ability of the model to recognize the intended class based solely on observed robot motion.  
For this test, we used the same physical platform as in the generation experiment, but reversed the interaction direction: the robot arm was manually moved along one of the 26 training trajectories, while the model passively observed the resulting end-effector positions in real time.
To simulate uncertainty in initial prior knowledge, the 26-dimensional class embedding $\mathbf{C}$ was initialized with $\mathcal{N}(0, 0.1^2)$ noise.
At each control step, the observed trajectory up to the current timestep was fed into the model, and the class embedding vector $\mathbf{C}$ was updated by minimizing the accumulated past prediction error, as described in Sec.~\ref{sec:past_reconstruction}.  
We used a learning rate $\alpha_c = 2.5 \times 10^{-2}$, a window size of $50$ timesteps and performed $n_{\text{iter}} = 15$ past-reconstruction iterations per timestep.
The computation time per timestep did not exceed approximately 130 ms.

Recognition was performed using the model that achieved the lowest DTW error in alphabet drawing experiment (i.e., the best-performing MultiLarge instance).  
Following each update, the model also predicted motion of the future timesteps using the updated $\mathbf{C}$, enabling real-time class inference and future prediction.

Each of the 26 motion classes was tested in 10 independent trials with different random initializations of $\mathbf{C}$, resulting in a total of \textbf{260 recognition trials}.  
At the end of each trial, the predicted class was defined as the index of the maximum value in $\mathbf{C}$ (Top-1), and the second-highest index was recorded as the Top-2 prediction.
Accuracy was evaluated across all trials using Top-1 and Top-2 classification rates.
In addition, the final reconstruction error of internal states was recorded for each trial, allowing to compare the error distributions across three outcome groups: Top-1 correct, Top-2 correct, and Incorrect.

\section{Result and Discussion}

\subsection{Training Result}
\label{sec:trainingResult}
As a baseline comparison, we begin by summarizing the training performance of all 60 CERNet networks (10 runs per model across 6 types).
Table~\ref{tab:minimum-loss-comparison} shows the average minimum prediction-error loss per model type, corresponding to the configurations listed in Table~\ref{tab:model-configs}.

Model performance improved with size across both architectures.
For example, among single-layer CERNets, Mini vs. Standard ($p = 1.1 \times 10^{-6}$) and Standard vs. Large ($p = 0.015$) were both significant.
Multi-layer variants showed the same trend.
Importantly, at every model size, the multi-layer version consistently outperformed its single-layer counterpart, with particularly strong effects in the Standard ($p = 3.8 \times 10^{-9}$) and Large ($p = 5.1 \times 10^{-13}$) settings.
Even in the Mini configuration, MultiMini showed a measurable advantage ($p = 0.0041$).
These results clearly indicate that hierarchical structure improves prediction accuracy, especially at higher capacities, but even small models benefit from layering.

\begin{table}[h]
    \centering
    \caption{Minimum Prediction Error Loss During Training ($\times 10^{-3}$)}
    \label{tab:minimum-loss-comparison}
    \begin{tabular}{lccc}
        \hline
         & Mini & Standard & Large \\
        \hline
        Single & A) $7.51 \pm 0.81$ & B) $4.60 \pm 0.91$ & C) $3.66 \pm 0.55$ \\
        Multi  & D) $4.53 \pm 2.59$ & E) $0.90 \pm 0.54$ & \textbf{F) \bm{$0.26 \pm 0.10$}} \\
        \hline
    \end{tabular}
\end{table}

\subsection{Alphabet Drawing Experiment}

In this section, we present the results of the alphabet drawing experiment in simulation and real-robot.
First, we show the Dynamic Time Warping (DTW) score to the ground-truth training trajectories of each model as a quantitative evaluation.
Next, we illustrate the actual alphabet drawings both in simulation and real-robot as a qualitative evaluation.

\subsubsection{Quantitative Evaluation: Dynamic Time Warping score}
Table~\ref{tab:DTW-score} summarizes DTW scores between generated and training trajectories.

\begin{table}[h]
    \centering
    \caption{DTW Score ($\times 10^{-1}$)}
    \label{tab:DTW-score}
    \begin{tabular}{lll}
        \hline
        Model ID  & DTW in Simulation  & DTW in Robot Exp.     \\
        \hline
        A) SingleMini & $1.70 \pm 0.84$    & -            \\
        B) SingleStandard & $1.23 \pm 0.78$  & $1.90 \pm 0.87$                \\
        C) SingleLarge & $1.03 \pm 0.70$   & $2.32 \pm 1.52$              \\
        D) MultiMini&  $1.07 \pm 0.73$  & -\\ 
        E) MultiStandard &  $0.44 \pm 0.28$ & $1.13 \pm 0.48$\\
        \textbf{F) MultiLarge} &  \bm{$0.25 \pm 0.11$} & \bm{$0.95 \pm 0.39$}\\
        \hline
    \end{tabular}
\end{table}

Table~\ref{tab:DTW-score} further illustrates the impact of hierarchical structure in simulation.
The three-layer \emph{MultiLarge} networks achieved the lowest DTW score ($0.25 \pm 0.11$), reducing alignment error by 76\% compared to \emph{SingleLarge} ($1.03 \pm 0.70$). 
Although differences at smaller scales were less pronounced (e.g., MultiMini vs.\ SingleMini), the multi-layer models consistently outperformed their single-layer counterparts at all sizes.

Additionally, while all models exhibit some performance degradation when transferred from simulation to the physical robot experiment, this effect is particularly visible in the single-layer architectures.
As shown in Table~\ref{tab:DTW-score}, the average DTW scores increase across all models, yet the scores alone do not fully capture the severity of the degradation in the single-layer models.
In particular, qualitative inspection of the reproduced motions (detailed in the next section) reveals that multi-layer models maintain legible trajectories that still resemble the intended characters, whereas single-layer models often fail to generate discernible shapes, despite their average DTW scores remaining within a similar range.

This discrepancy highlights the limitation of purely quantitative metrics in capturing perceptual fidelity and further supports the practical advantage of hierarchical architectures in real-world deployment.

\subsubsection{Qualitative Analysis: Actual Alphabet Drawing}

\begin{figure}[htbp]
  \centering
  \includegraphics[width=1\linewidth]{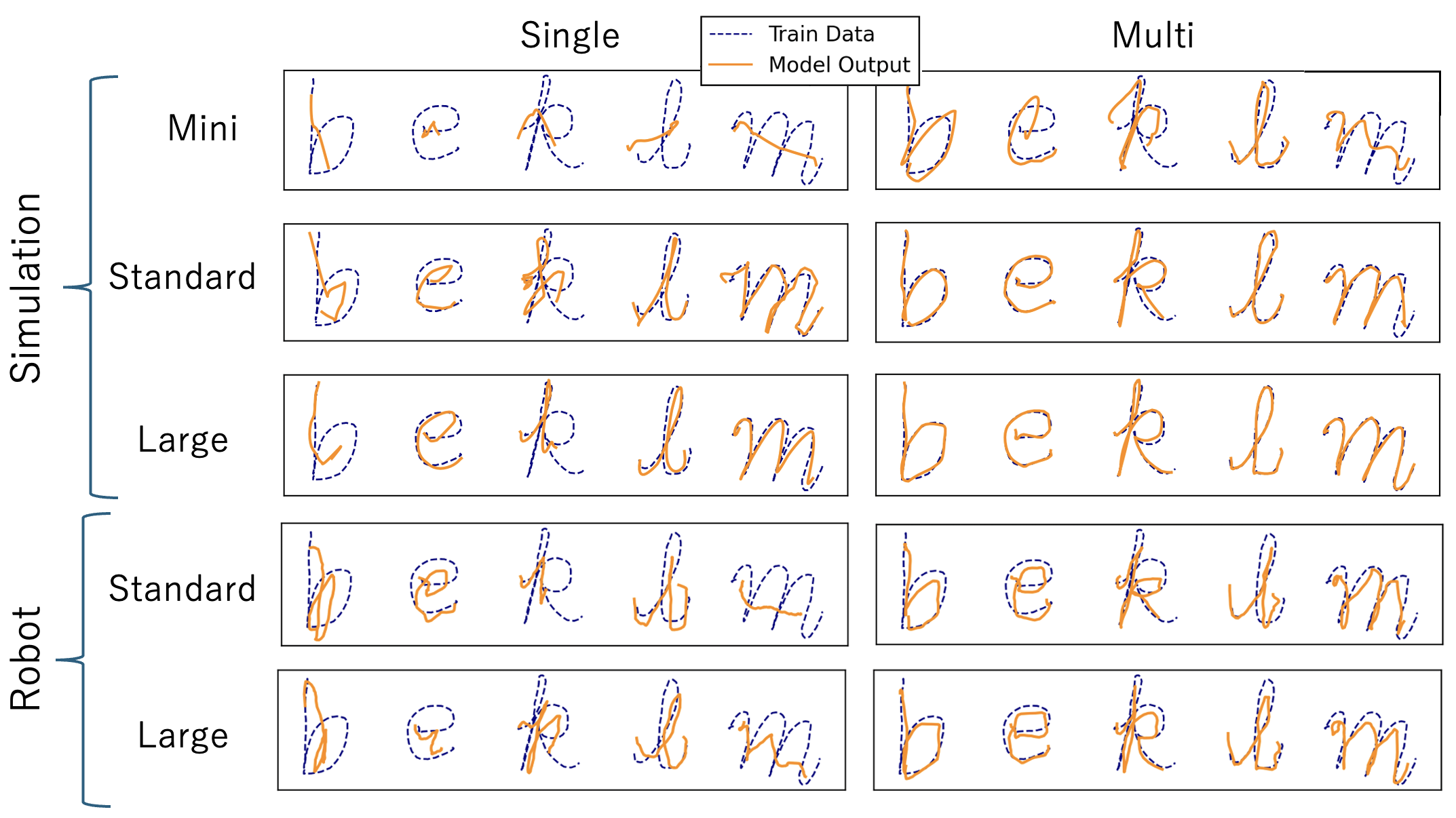}
  \caption{Reproduction of the letters \emph{b, e, k, l, m} by the best-performing networks of each model type. Each row corresponds to a different model scale and experimental condition (simulation or real robot), while columns compare single-layer and multi-layer architectures. Dotted lines indicate the original training trajectories, and solid lines represent the generated motions by the models.}
  \label{fig:alphabetAGeneration}
\end{figure}

To qualitatively evaluate the legibility of reproduced trajectories, we selected five representative characters (b, e, k, l, m) and visualized the motions generated by the best-performing instance of each model.
The results are shown in Fig.~\ref{fig:alphabetAGeneration}, illustrating how model architecture affects stroke structure and spatial coherence.

Multi-layered CERNets produce visually recognizable characters that align well with the training data, consistent with their low DTW scores.
In contrast, single-layer models show greater distortion, especially in complex characters like 'k' and 'b', reflecting their higher DTW scores.
Despite having fewer parameters, MultiMini (D) matches SingleLarge (C) in output quality, highlighting the efficiency of hierarchical structures.

When transferred to the real robot, all models exhibit some performance degradation. However, while single-layer models often produce illegible outputs, multi-layer models still retain clearly recognizable character shapes. This further supports the practical robustness of hierarchical architectures in noisy, real-world settings.
These observations imply that hierarchical architectures improve not only numerical alignment with training data, but also the qualitative correctness of generated motion.

\subsection{Perturbation Resistance Experiment}

\begin{figure}[htbp]
  \centering
  \includegraphics[width=0.75\linewidth]{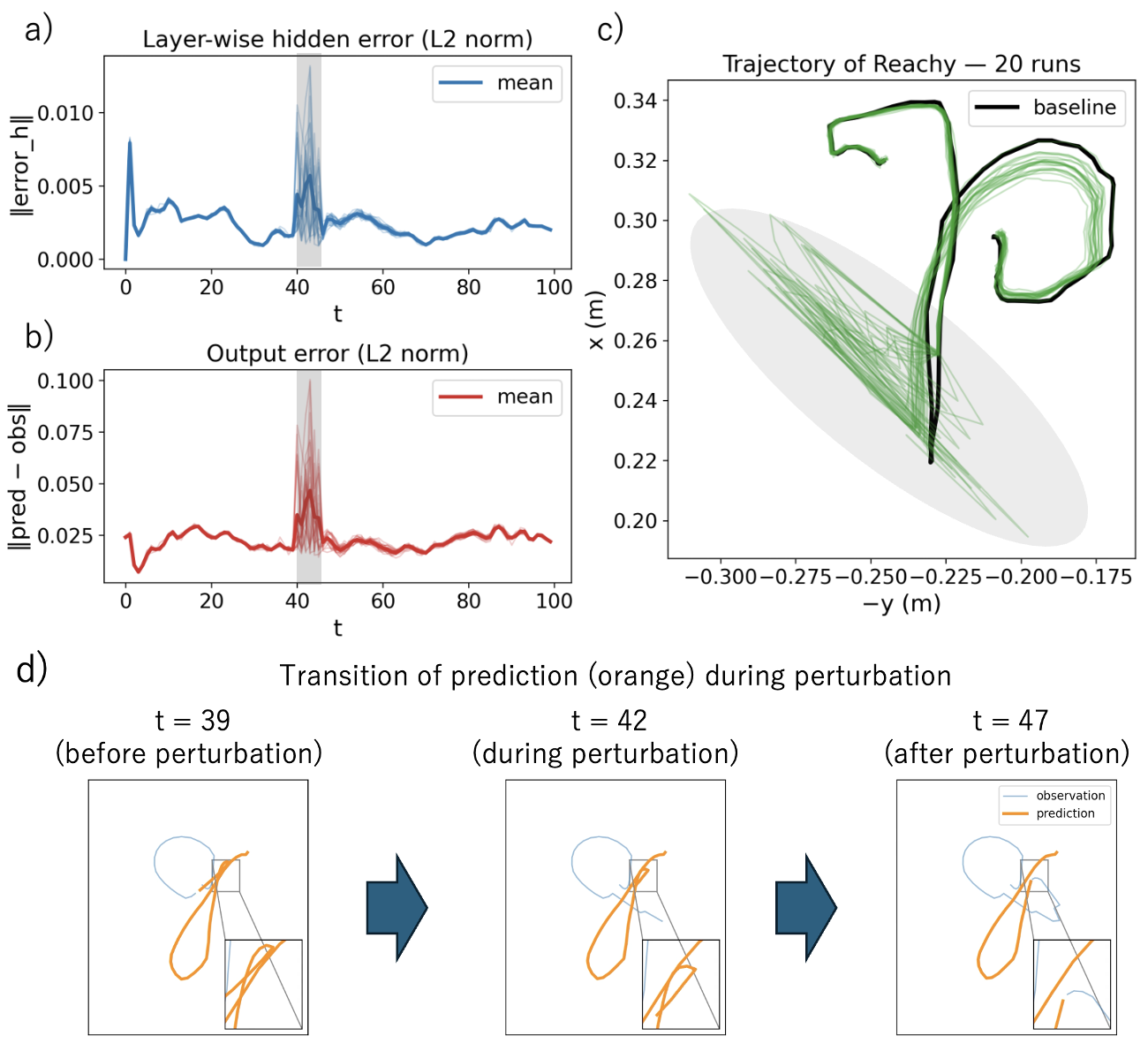}
  \caption{Perturbation recovery during alphabet reproduction on Reachy using multi-layered CERNet model. (a) and (b): Layer-wise prediction errors with respect to time while reproducing letter \texttt{p}. The grey area indicates the timesteps where perturbation was injected. (c) End-effector trajectory while reproducing letter \texttt{p}. (d) Future trajectory prediction while writing \texttt{g} under an injected perturbation: future prediction of the trajectory realigns as internal states adapt.}
  \label{fig:perturbation}
\end{figure}

In this section, we present the result of the perturbation experiment conducted on Reachy using the proposed multi-layered CERNets. 
A demonstration video of a representative trial is available,\footnote{Video available at: \url{https://git.cyu.fr/hiroki/cernet_paperresultreproduction/-/blob/main/perturbation_youtube.mp4}} 
which shows the robot drawing the letter \texttt{p} while an external disturbance is applied between timestep 40 and 45 (see Fig.~\ref{fig:perturbation}).
Fig.~\ref{fig:perturbation} shows a representative trial where the robot was drawing the letter \texttt{p} and an external disturbance was applied between timestep 40 and 45.

Fig.~\ref{fig:perturbation} a) and b) reveal that the prediction error, used to update internal states, increased sharply during the perturbation period, indicating that the model identified the deviation between prediction and observation.  
Fig.~\ref{fig:perturbation} c) demonstrates that, once the external disturbance ceased, the robot’s trajectory gradually converged back to the original path, confirming the self-correcting capability of the model.
Moreover, as shown in Fig.~\ref{fig:perturbation} d) (for a different trial drawing the letter \texttt{g}), the predicted trajectory after the disturbance gradually shifted, illustrating how future predictions are updated through internal error minimization.

Overall, this experiment highlights the ability of CERNet to withstand external disturbances by dynamically adjusting its internal states, spontaneously recovering the learnt trajectory.

\subsection{Class Inference \& Confidence Estimation Experiment}

In this section, we present the result of the class inference and confidence-level estimation experiment conducted on Reachy using multi-layered CERNet model.
A demonstration video of a representative trial is available,\footnote{Video available at: \url{https://git.cyu.fr/hiroki/cernet_paperresultreproduction/-/blob/main/Experiment3_InferLetter.mp4}} 
which illustrates how the model infers the intended character "b" in real time given the trajectory.
The goal of this experiment is: (i) to evaluate the model’s ability to infer the intended trajectory class in real time from observed motions, and (ii) to assess whether the internal prediction error reflects the model’s confidence in its own inference.

To evaluate class inference, we analyzed the 260 physical trials described in the experiment section, in which Reachy's arm was manually moved along one of the 26 training trajectories while the model observed the end-effector positions.
The class embedding vector $\mathbf{C}$ was updated online at each timestep by minimizing past prediction error, and the final inferred class was determined by the index with the highest value in $\mathbf{C}$ (Top-1), with the second-highest taken as Top-2.

Across 260 trials (10 trials per class), the model achieved a Top-1 classification accuracy of \textbf{68\%} and a Top-2 accuracy of \textbf{81\%}. 
These results demonstrate that the model is capable of performing real-time recognition over a large set of motion classes in a noisy physical environment.

To evaluate the model’s implicit confidence estimation, we compared the final mean squared error (MSE) of the past reconstruction loss $L_\mathrm{past}$ among three groups: correctly recognized at Top-1, correctly recognized at Top-2, and incorrect.
The results are summarized below:

To assess whether the model's internal reconstruction error reflects recognition accuracy, we compared the final past reconstruction MSE across three categories: correctly recognized at Top-1, recognized at Top-2, and incorrect predictions.  
Statistical comparisons using the Mann–Whitney U test revealed that the MSE in the Top-1 group was significantly lower than that in the Top-2 group ($U = 1723.0$, $p = 0.00019$), and also significantly lower than the Incorrect group ($U = 2038.0$, $p < 10^{-8}$).  
Although the Top-2 group tended to show lower error than the Incorrect group, the difference was not statistically significant ($U = 622.0$, $p = 0.060$).  
It should be noted that the variance of the errors within each group was relatively large, suggesting that further analyses will be required to fully characterize the relationship between error magnitude and confidence.
Overall, these results suggest that the model’s internal prediction error serves as an implicit confidence measure: the lower the error, the more likely the inferred class is correct.

These findings suggest that the model is capable of internally differentiating between correct and incorrect inferences based on its own prediction error.
In particular, lower MSE is correlated with more accurate recognition, implying an emergent internal confidence indicator.

\begin{figure}[htbp]
  \centering
  \includegraphics[width=0.95\linewidth]{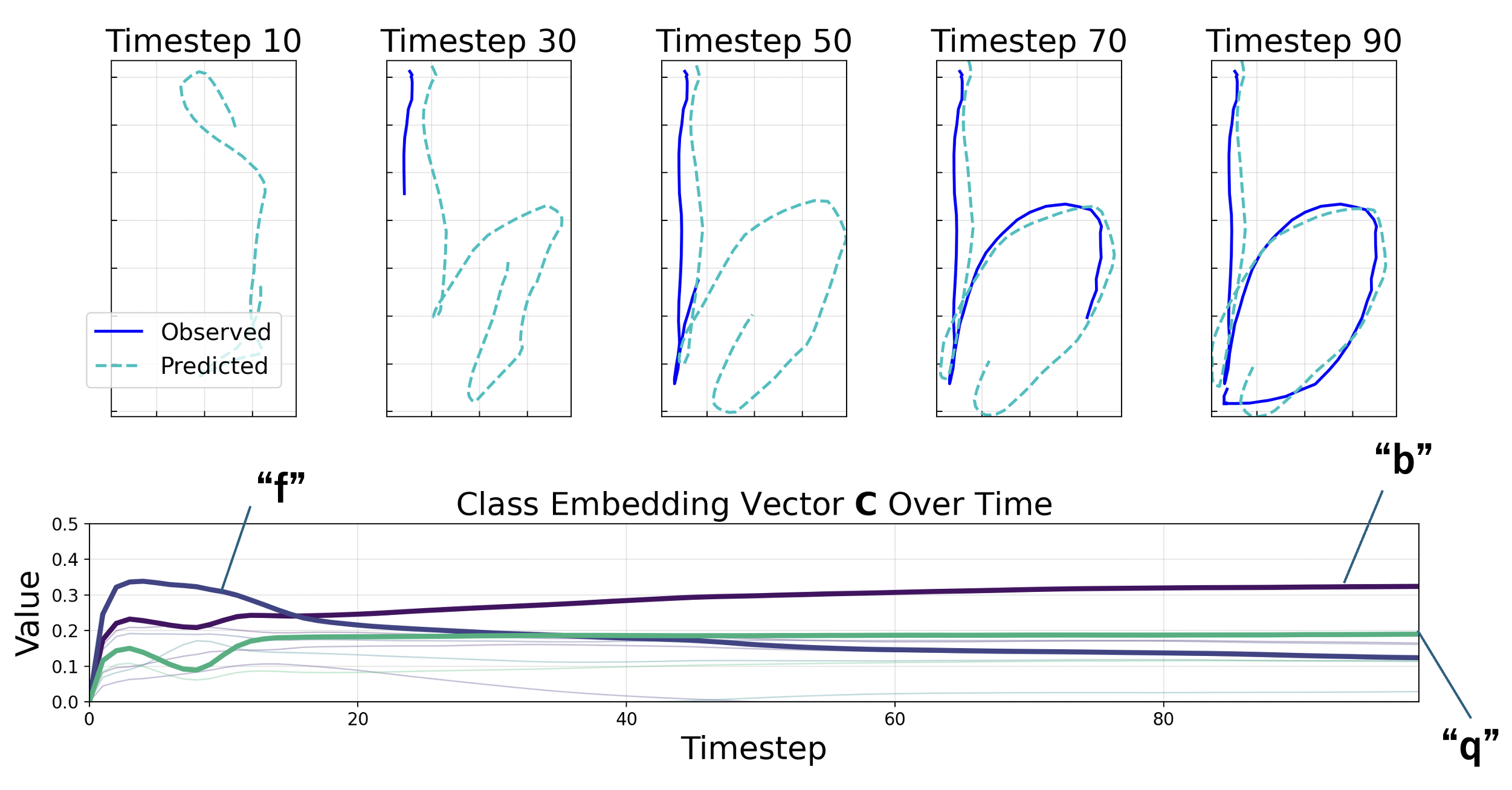}
  \caption{Time development of the class embedding vector of CERNet in \emph{inference mode}. The top row illustrates the time development of the prediction and the observation from the robot. Whereas the bottom row illustrates the time development of the intrinsic prediction of the observed motion.}
  \label{fig:classInference}
\end{figure}

\begin{figure}[htbp]
  \centering
  \includegraphics[width=0.85\linewidth]{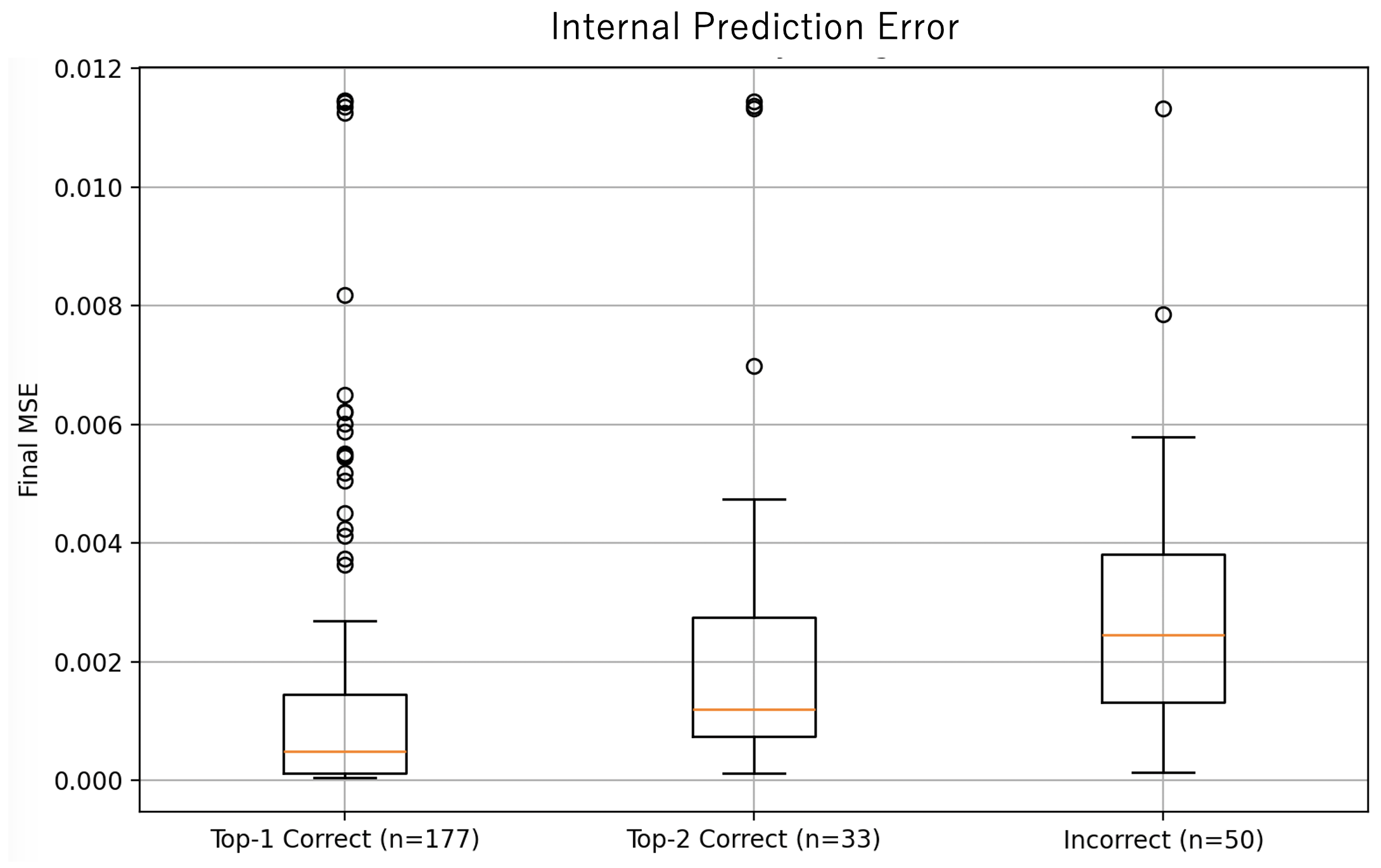}
  \caption{Final mean squared error (MSE) of the internal prediction error, averaged over time, grouped by recognition outcome.
Each boxplot shows the distribution of reconstruction error across trials where the correct class was identified at Top-1 (left), at Top-2 (middle), or was not among the top two predictions (right).
Lower errors correspond to higher inference accuracy, indicating that the model's prediction error implicitly reflects its confidence.}
  \label{fig:mseInClassInference}
\end{figure}

As an illustrative example, Fig.~\ref{fig:classInference} shows a representative trial in which the model infers the character "b".
As time progresses, the updated class embedding gradually drifts toward the subspace corresponding to "b", and both past reconstruction and prediction align more closely with the ground-truth trajectory.
Furthermore, Fig.~\ref{fig:mseInClassInference} displays a boxplot of the final past reconstruction MSE for each recognition category.
The trend confirms that the model exhibits higher prediction error when its recognition is incorrect, and lower error when it is confident and correct, supporting our claim of self-monitoring capability.

\section{Conclusion}

This study presented a unified neural architecture, CERNet, Class-Embedding Predictive-Coding RNN, that integrates recognition, generation, and confidence estimation within a single model and operates in real time on a physical humanoid robot.

By combining a multi-layer predictive-coding framework with a shared class embedding vector, the proposed model enables scalable learning and regeneration of complex motor repertoires while minimizing interference between patterns.
We demonstrated that this architecture successfully stores and reproduces 26 handwritten alphabet trajectories on the humanoid robot, Reachy, achieving a 76\% reduction in reproduction error compared to a single-layer baseline with almost an identical number of parameters.
The hierarchical structure not only demonstrated accuracy but also enhanced robustness, as shown in Experiment 2 by spontaneous recovery from physical perturbations through online error correction.

Furthermore, we showed that the same model can infer the demonstrated motion class in real time, achieving 68\% Top-1 and 81\% Top-2 classification accuracy without retraining.
Importantly, the past-reconstruction error serves as an indicator of confidence: trials with correct recognition exhibited significantly lower past-reconstruction error than incorrect ones, suggesting that the model can internally assess the certainty of its inference.

Together, these findings position CERNet as a lightweight, real-time-capable control model that integrates trajectory generation, class recognition, and confidence estimation within a single framework.
Its ability to encode a large repertoire of motion patterns, maintain robustness to perturbations through internal state adaptation, and infer motion class with measurable confidence makes it particularly well-suited for embodied interaction scenarios.

We believe this architecture provides a promising foundation for future research in human–robot interaction, where smooth and natural collaboration requires real-time understanding of intent, adaptive behavior execution, and intrinsic evaluation of inference reliability.
Future work will explore its extension to online learning and integration with multimodal sensory inputs such as vision, enabling more complex and context-aware interaction behaviors.

\bibliographystyle{IEEEtran}
\bibliography{references}

\end{document}